\documentclass{article}

\usepackage[final]{neurips_2023}

\usepackage[utf8]{inputenc} 
\usepackage[T1]{fontenc}    
\usepackage{hyperref}       
\usepackage{url}            
\usepackage{booktabs}       
\usepackage{amsfonts}       
\usepackage{nicefrac}       
\usepackage{microtype}      
\usepackage{xcolor}         
\usepackage{biblatex}
\usepackage{pifont}
\newcommand{\cmark}{\ding{51}}%
\newcommand{\xmark}{\ding{55}}%
\usepackage{graphicx}

\title{Human Shape and Clothing Estimation}

\author{%
  Aayush Gupta* \\
  Department of Computer Science\\
  University of California, San Diego\\
  San Diego, CA 92092 \\
  \texttt{aag011@ucsd.edu} \\
  \And
  Aditya Gulati* \\
  Department of Computer Science\\
  University of California, San Diego\\
  San Diego, CA 92092 \\
  \texttt{adgulati@ucsd.edu} \\
  \And
  Himanshu* \\
  Department of Computer Science\\
  University of California, San Diego\\
  San Diego, CA 92092 \\
  \texttt{hhimanshu@ucsd.edu} \\
  \And
  Lakshya LNU* \\
  Department of Computer Science\\
  University of California, San Diego\\
  San Diego, CA 92092 \\
  \texttt{llnu@ucsd.edu} \\
}

\begin{document}
\maketitle
\begin{abstract}
Human shape and clothing estimation has gained significant prominence in various domains, including online shopping, fashion retail, augmented reality (AR), virtual reality (VR), and gaming. The visual representation of human shape and clothing has become a focal point for computer vision researchers in recent years. This paper presents a comprehensive survey of the major works in the field, focusing on four key aspects: human shape estimation, fashion generation, landmark detection, and attribute recognition. For each of these tasks, the survey paper examines recent advancements, discusses their strengths and limitations, and qualitative differences in approaches and outcomes. By exploring the latest developments in human shape and clothing estimation, this survey aims to provide a comprehensive understanding of the field and inspire future research in this rapidly evolving domain.
\end{abstract}

\section{Introduction}
Human shape and clothing estimation is a complex and challenging task with a wide range of applications across various domains. Accurate estimation of human shape and clothing has numerous practical applications. In the realm of fashion retail, virtual try-on systems \cite{ghodhbani2022you} empower customers to visualize how different garments would appear on their bodies without physically trying them on. This technology enhances the online shopping experience, reduces returns, and improves customer satisfaction. Moreover, personalized fashion recommendations based on individual body shapes and style preferences can be facilitated. The field of virtual reality (VR) and augmented reality (AR) also greatly benefit from human shape and clothing estimation \cite{liu2020comparing}. By accurately replicating users' physical appearances and clothing choices, immersive virtual environments can offer a more personalized and engaging experience. Applications include gaming, social VR, virtual meetings, and virtual fashion shows. In animation and visual effects, precise estimation of human shape and clothing is crucial for creating realistic characters and enhancing the visual quality of films, video games, and visual effects. Accurate modeling and rendering of body shape and clothing details contribute to a visually appealing and immersive experience.

However, the task of human shape and clothing estimation is non-trivial due to several challenges. Variations in body shapes, sizes, and poses present difficulties in accurately estimating body shape and proportions. The diversity of clothing styles, textures, and patterns adds further complexity to the task. Occlusions caused by overlapping clothing layers, objects, or body parts, along with variations in image quality, lighting conditions, and camera viewpoints, further hinder accurate estimation. Moreover, obtaining a large amount of labeled training data with precise annotations poses a challenge for building accurate and robust estimation models.

Several prior works have contributed to the understanding and advancement of clothing analysis techniques and clothing image retrieval technology. In a survey by Liu et al. \cite{liu2014fashion}, a comprehensive overview of clothing analysis techniques is presented, including clothing modeling, retrieval, and recommendations. The survey also delves into makeover analysis, exploring topics such as facial makeup, hair beauty, and related areas. Another significant work by Ning et al. \cite{ning2022survey} focuses specifically on clothing image retrieval technology, which has gained substantial popularity in e-commerce platforms and search fields, such as Taobao, Jingdong, Baidu, Google, among others. The authors primarily investigate clothing image retrieval in cross-domain situations, where the image to be queried and the image retrieval database originate from two distinct domains.

In this survey paper, we present an overview of the field of human shape and clothing estimation with the objective of exploring and analyzing the existing research in this area, focusing on the background, methods, and findings related to human shape estimation, fashion generation, landmark detection, and attribute recognition.

The paper is structured as follows: Section \ref{sec2} provides a detailed explanation of the background, highlighting the critical requisite knowledge of human shape estimation and clothing estimation in various domains. It discusses the non-trivial nature of the tasks, and the motivation to solve the clothing estimation problem. In Section \ref{sec3}, we delve into the methods employed for human shape estimation, fashion generation, landmark detection, and attribute recognition. We perform the critical analysis in this section. The section also covers different approaches, algorithms, and models utilized to accurately estimate human shape, generate realistic virtual clothing, detect landmarks on the human body, and recognize clothing attributes. We discuss these methods in terms of strengths, motivations, underlying reasons, limitations, and qualitative differences in approaches and outcomes. Finally, in Section \ref{sec4}, we conclude the paper by summarizing our findings and insights regarding human shape and clothing estimation. We highlight the key takeaways from the surveyed literature, identify the current challenges and future research directions, and discuss the potential applications and implications of the advancements in this field.

Through this survey paper, we aim to provide a comprehensive understanding of the state-of-the-art techniques and developments in human shape and clothing estimation. We hope that the insights and knowledge gained from this survey will contribute to the ongoing research in this field and inspire new avenues of exploration and innovation.

\section{Background}\label{sec2}
Before we explore the latest methods for estimating human body shape and clothing, we describe how we can represent the human body using shape and pose parameters. We give an overview of the computer vision tasks involved in estimating human shapes and clothing in the field of fashion technology in the subsequent subsection.

\subsection{Human Body 3D Representation}
Understanding and accurately modeling the human body has been a persistent challenge in computer graphics, animation, virtual reality, and bio-mechanics \cite{10.1145/2816795.2818013}. Estimating human body pose and shape from 2D images is particularly complex due to the loss of information from the 3D space. The highly articulated nature of the human body, along with the dimensional complexity, unusual and complex poses, and contrast variations introduced by clothing, further complicate the modeling difficulties. 

Numerous research works \cite{inproceedings, 216727, 895978, GAVRILA199982} have explored approaches to accurately capture the human body shape using 3D models. These methods often rely on incorporating priors such as contours, silhouettes, cylindrical representations, or meta-balls models. However, these earlier techniques encountered significant challenges, resulting in ill-posed, and unrealistic representations of the human body. The fundamental issue with these approaches was the inability to achieve a precise and reliable model that accurately reflected the complexities of the human form. The resulting models often suffered from mismatches, lacked robustness, and failed to capture the true realism of the human body. This limitation hindered the progress in achieving more realistic and versatile representations, preventing the applications of these models from reaching their full potential.

Skinned Multi-Person Linear Model (SMPL) \cite{10.1145/2816795.2818013} is a vertex-based parametric model to represent human body shapes accurately. SMPL is a generative model in which there are two key components namely \textit{shape} $\beta$ and \textit{pose} $\theta$. The \textit{shape} parameter represents the variation in height, weight, body proportions, contours and the \textit{pose} parameter represents the 3D surface deformation with articulation. SMPL presents a differentiable function $M(\theta,\beta) \in R^{3 \times N}$ - which outputs a triangulated mesh with N = 6980 vertices. Given the body mesh, the body joints X can be modeled as linear combination of vertices, hence a pre-trained linear regression W is defined for k joints of interest to derive X $X \in R^{k \times 3} = WM$.

Given this parametric model, the human shape estimation papers described in subsequent sections are based on estimating the aforementioned shape and pose parameters. The main pivot of all the methods is training these parameters by extracting features from single/multiple RGB images. These methods deploy various strategies like adversarial training \cite{kanazawa2018endtoend}, iterative model fitting \cite{kolotouros2019learning}, pressure imaging \cite{clever2020bodies}, neural mesh rendering \cite{Yang2022}, synthetic data augmentation \cite{sengupta2020synthetic},  to improve the training, and model the shape parameters accurately.  Another significant class of body estimation methods involves Body Mesh points \cite{jrna}, pixel aligned implicit functions \cite{pifu,saito2020pifuhd}, volumetric performance capture  and novel-view rendering \cite{li2020monocular}. More details of these methods with critical analysis are described in the subsequent sections. 

\subsection{Intelligent Fashion}
Due to the value of the fashion industry, a lot of research areas have been developed related to intelligent fashion. By the term intelligent fashion, we mean such methodologies that focus on solving fashion-related tasks using machine learning. 

Fashion detection is one such broad category of tasks. Many of the fashion-related solutions utilize detection methods as a given for their method. For example, a landmark detection task \cite{liu2016deepfashion} might be used beforehand by a recommendation system. Landmark detection aims to predict the positions of key points such as neckline, hemline, cuff, etc on clothing. Other tasks that might also fall under this category are fashion parsing \cite{yamaguchi2012parsing}, wherein a segmentation mask for different types of clothing items like pants or dresses is generated, and image-based fashion item retrieval.

Fashion no longer refers to just the appearance of the person, but is also linked to the personality traits and various social cues. Thus, analyzing fashion for precision marketing or for sociological reasons is important. Similarly, categorizing clothing based on various attributes such as category, pattern, neckline shape, formal or informal, can be very helpful for recommendation-based tasks or for quantifying fashion style. Although, fashion style can also be represented by directly learning discriminative features \cite{kiapour2014hipster} through machine learning methods for different styles like bohemian, pinup, goth, etc., and use them to understand various trends. Popularity prediction and fashion forecasting is another important analysis task, not just for marketing campaigns but also for recommendations to individuals on specific occasions. While popularity prediction is person agnostic, fashion recommendations can also be made for a particular individual or clothing item, such as fashion compatibility recommendation \cite{iwata2011fashion}, wherein the task is to understand the compatibility between clothing items of various categories.

We can also perform synthesis-related tasks apart from detection and analysis-type tasks. Synthesis-related tasks can greatly reduce the effort of users to have to manually try different makeups, clothing items, etc, and thus are very valuable for the e-commerce industry. Style transfer tasks for instance such as facial make-up try-on \cite{li2015simulating} or virtual clothing try-on \cite{han2018viton}, can greatly reduce the cost of having to order multiple candidates online and thus save transportation costs. Style transfer tasks also need to allow pose transformation flexibility for a truly in-store experience.

It is apparent from the previous paragraphs that intelligent fashion methods are trying to accomplish highly diverse tasks, resulting in the development of varied specialized techniques related to a task. Thus, in this survey paper, we restrict our focus to predominantly three fashion tasks, Fashion Generation, Landmark Detection, and Attribute Recognition. We take a deeper look into the problem formulation, as well as various methods that have been developed over the years for these tasks in further sections.

\section{Methods and Critical Analysis}\label{sec3}

This section covers a detailed description of state-of-the-art methods in human shape estimation, which acts as a foundation for the intelligent fashion areas - Clothing Generation, Landmark Detection, and Attribute Recognition. The section also covers the analysis of these methods in terms of strengths, motivations, underlying reasons, limitations, and qualitative differences in approaches and outcomes.

\subsection{Human Shape Estimation}
The human shape estimation from a single RGB image is a challenging task in computer vision. It has several applications in the field of HCI, graphics, virtual try-on, action recognition, etc. The primary paradigm of human shape estimation as described in the background section is estimating the shape and pose parameters for the SMPL - a generative parametric model to estimate the mesh vertices. The methods described below utilize this SMPL model in some way.

The HMR (Human Mesh Recovery) \cite{kanazawa2018endtoend} stands as a significant early contribution to the field of estimating human shape and pose. It introduces an end-to-end framework that enables the reconstruction of a 3D mesh representing the human body from a single RGB image. The method involves passing the input image through a CNN encoder to extract relevant representations. These representations are then fed into an iterative regressor, which learns the camera, pose, and shape parameters. These parameters serve as input to the SMPL model, allowing for the generation of a 3D mesh of the human body.

The method then proceeds to generate keypoints that can be compared with the ground truth keypoints. This comparison is utilized to calculate the reprojection error, which quantifies the accuracy of the reconstructed mesh. The paper's main contribution lies in the utilization of an unpaired 3D dataset to train an adversary, referred to as a discriminator. The discriminator's role is to differentiate between natural parameters and the generated parameters,  ensuring that the generated parameters are realistic to model human body shapes.

The main strength of HMR is that it trains the whole method in an end-to-end manner, hence resulting in better accuracy and low run time. Moreover, HMR opens up possibilities of utilizing unpaired 2D-to-3D data in an adversarial manner which is available in more abundance. Although the HMR is considered as a good performing benchmark for many of the recent papers in human shape estimation it has showcased various limitations in terms of high sensitivity to image quality and clothing variations, poor performance in complex poses, and dependency on high amounts of training data.

HMR is a regression-based method to directly estimate model parameters from RGB images, as mentioned such methods require huge amounts of supervision. Other class of methods includes classical optimization-based methods to iteratively fit SMPL-based models to 2D observations. These methods give better accuracy with lesser data (unlike HMR) but are slow and sensitive to initialization. SPIN \cite{kolotouros2019learning} proposes an approach which merges both the paradigms uniquely to estimate human shape. It utilizes the same regression architecture as that of HMR but the inferred output acts as an initialization to the optimization method SMPLify \cite{SMPL-X:2019}. Along with the reprojection error as described in HMR, the SPIN paper regularizes the loss with a parametric loss between the regressed parameters $\Theta_{reg}$ and iterative fitting parameters $\Theta_{opt}$. 

The main characteristic strength of this method is that it is self-improving in nature, good initialization from regressor leads to better fit and a good fit can give better supervision. Hence the optimization-module and regression-module form a self-improving symbiotic cycle. The paper still fails to complex challenging poses, and is prone to erroneous reconstructions due to ordinal depth ambiguities, viewpoints rarely occurring in the dataset.

Complex poses and occlusion has posed as a serious challenge to the above two methods, PressureNet \cite{clever2020bodies} has been one of the earliest works to tackle the complex human poses limited to body at rest position. PressureNet utilizes the generative part of the HMR network to output the differentiable mesh and proposes a novel architecture to incorporate pressure map reconstruction network that models pressure image generation to enforce consistency between estimated 3D body models and pressure image input. The paper opens up the possibility of tackling complex poses but it is still limited to the niche involving physics simulations to generate pressure images.

PressureNet addresses the occlusion problem, but its applicability is limited to a specific domain. Recognizing this limitation, Yang et al. (2022) were motivated to develop a novel approach for accurately learning human pose and shape using occlusion-aware data. The proposed method focuses on categorizing different types of occlusions, including self-occlusion, object-human occlusions, and inter-person occlusions. Their framework introduces a distinctive approach by synthesizing occlusion-aware silhouettes and 2D keypoints and directly regressing them to SMPL pose and shape parameters. A notable aspect of this paper is the utilization of a neural 3D mesh renderer, enabling real-time supervision of silhouettes. The primary objective of the paper is to address the scarcity of data in scenarios involving occlusions. However, LASOR still poses a requirement of high training time and is bottle-necked by the performance of the synthetic occlusion data generator.

In summary, all of these works revolve around the central concept of estimating SMPL parameters from RGB images. The primary distinction among these methods lies in their strategies for incorporating additional sources of supervision to generate realistic predictions for these parameters. The HMR method serves as a foundational model that has paved the way for the subsequent methods mentioned earlier. Accurate human shape estimation serves as a fundamental prerequisite for various intelligent fashion tasks, as discussed in the following subsections. Acquiring a deeper understanding of these shape estimation methods facilitates a better comprehension of the diverse tasks at hand.

\subsection{Fashion Generation}
The virtual try-on task involves creating a computer-generated simulation of trying on clothing items or accessories. It aims to generate a virtual representation of a person, $I_t$, and accurately render the chosen clothing item, $C$, onto that virtual person, $I_S$, allowing users to visualize how the clothing looks and fits without physically trying it on. The task combines computer vision and computer graphics techniques to estimate the pose and shape of the person, deform the clothing model accordingly, and render the virtual try-on result, providing a convenient and immersive virtual shopping experience. 

One of the early works using DL for Virtual Try-On was present in the VITON\cite{han2018viton} paper. The paper presented a two-stage model for the task. The first method would take in a person representation and the target clothing $C$ to generate coarse results $I'$ and a clothing mask $M$. Then, a thin plate split (TPS) transformation, whose parameters are evaluated using shape-aware context matching, would be applied to the clothing $C$ to get warped $C'$. Finally, in the second stage model would predict a composition mask $\alpha$, such that the final output would be $I_t = \alpha C' + (1-\alpha) I'$. Perceptual loss is used in both stages, along with L1 loss in the first stage and smoothing loss in the form of total variation regularizer in the second stage.

An important part of the method is the person representation used in stage-1. Typically, some variation of this representation is used in 2D virtual-try-on models. A multi-channel input is prepared of the same height and width as the reference image $I$ containing 
\begin{itemize}
    \item An-18 channel pose heatmap generated using SOTA pose estimator, where each channel represents one of the 18 key points.
    \item A 4-channel human segmentation map using SOTA human parser. The first channel in the map represents the binary mask representing the human shape, while the other 3 channels provide face and hair segmentations from the $I$.
\end{itemize}
Notably, the reference image $I$ is not used thus making representation clothing agnostic.

However, since VITON uses a perceptual loss with a coarse prediction as well, the second stage would be biased towards coarse prediction to improve upon the perceptual loss for the refined image. Thus, leading to poor performance on clothing details transfer. CP-VTON\cite{wang2018characteristicpreserving} improved upon the VITON's ability to preserve clothing details by removing the coarse prediction. Instead, the first stage predicts the TPS parameters, which are used to warp the cloth $C$ to $C'$. This stage uses L1 loss between $C'$ and the ground truth warped cloth for training. In the second stage, a UNet model then predicts a composition mask $\alpha$ and an initial prediction $I'$. The final output is again, $I_t = \alpha C' + (1-\alpha) I'$. Note, herein the initial prediction $I'$ can be thought of as a coarse prediction but all the losses are defined in terms of the final composited output. CP-VTON also uses the same cloth-agnostic person representation as VITON.

Followup work, CPVTON+\cite{Minar_CPP_2020_CVPR_Workshops}, made several small changes in the CPVTON to get sharper and better-textured results. In the first stage, they provided a complete target body silhouette area as input allowing improvement in hair occlusion artifacts. Also, to control distortion artifacts in warped clothing, they added a grid deformation regularization to the loss. Again, for the second stage, they added face, hair, legs, and clothes to the human representation, and also provided a binary mask of warped clothing as input, to prevent confusion of warped clothing with the background.

One of the major limitations of the works that we have described so far is that the TPS transformation used to warp clothing $C$ has a limited degree of freedom(2x5x5). A more flexible approach was proposed by ClothFlow\cite{clothflow}, wherein a dense flow is learned given the source clothing and the target segmentation map. It is important to note that Clothflow can perform both pose-guided person image generation as well as virtual try-on with a slight modification. To perform pose-guided person image generation ClothFlow also involves an initial target segmentation mask, $S_t$, generation stage, before performing cloth deformation(Stage-2) and final rendering(Stage-3). Both initial $S_t$ generation and final rendering of $I_t$ utilize a UNet-based architecture with appropriate inputs. The initial $S_t$ generation uses a Source Image $I_s$, source seg $S_s$ and the target pose $p_t$ to generate $S_t$, while the final rendering utilizes warped cloth $C'$, source image $I_s$, target segmentation mask $S_t$ and target pose $p_t$ to generate $I_t$. The second stage for cloth deformation utilized two feature pyramid networks one for the source elements, source cloth $C$, and source segmentation $S_s$, and the other for target segmentation mask $S_t$ predicted in the first stage. Taking inspiration from the optical flow literature, authors used cascaded coarse to fine flow estimation to predict the final warping flow. This paper also used style loss on the final render for better learning of texture details. To use this method for the virtual try-on task, the provided cloth image $C$ is used as the source image, while the target pose is kept as the source person's pose. 

Authors of \cite{sdafn} proposed a single-stage virtual try-on method as compared to previous methods discussed here, wherein no loss is introduced between warped clothing and ground truth clothing. More importantly, they introduced the concept of Multi-flow Estimation for structure information. Arguably, structural information is associated with multiple locations in the feature maps, and thus a single flow is not enough to estimate the structure information correctly. To achieve the same, they take inspiration from Deformable Attention\cite{zhu2021deformable} and develop Deformable Attention Flow Network(DAFN), wherein given input source and reference features, $x_s$ and $x_r$ respectively, the flow, $o$, and attention maps, $a$, are generated. Notably, $o \in \mathcal{R}^{2K \times W \times H}$ and $a \in \mathcal{R}^{K \times W \times H}$, where $K$ is the number of samples in deformable attention. Internally, DAFN computes self and cross-deformable attentions between source and reference features.

A separate line of work involves using 3D modeling for performing the task of virtual try-on. Specifically, CloTHVTON+ \cite{clothvton} argued that for complex poses, cloth deformation in 3D is much more quality-preserving and leads to natural deformation. Thus, they first construct a 3D clothing model through a reference SMPL model which has a far simpler shape and pose (A-pose) than the target person. Then, using SMPLify-X \cite{SMPL-X:2019} 3D body model of the input person is generated. The 3D clothing model is then deformed to match the 3D body model to get a deformed/warped 3D clothing model. Similar to the previous approach, a target segmentation mask is also initially generated by the approach. However, for the final image generation, two different models were used. A Parts Generation Network(PGN), a UNet, is used to generate the target body skin parts. Finally, a Try-on Fusion network(TFN), another UNet, fuses target human segmentation, warped cloth, and human skin parts together to generate the final output.

\subsubsection{Summary} Table-\ref{table:vton_method} gives a succinct comparison of the approaches. It can be seen that all virtual try-on methods have two common steps cloth deformation step to generate an intermediate output $C'$, and a final generation step. Methods like SDAFN\cite{sdafn} perform both of these tasks in a single-stage manner with only final image-based supervision, while other methods usually utilize two different stages to perform these tasks. Additionally, sometimes models use an additional stage to first generate a target segmentation map, which has been argued to perform better than a coarse body silhouette. Also, usually, a UNet model is used for final image generation, wherein some older methods also generate an $\alpha$ composition mask, which is used to combine an initial prediction with the warped clothing $C'$. Finally, the most important and varied aspect of these methods is the cloth deformation step. While earlier approaches have relied upon methods like TPS, newer approaches generally either rely on a 3D model for clothing deformation or learn a dense flow, even multiple flows, to perform clothing deformation. Techniques that learn a dense flow often also utilize a cascaded coarse-to-fine estimation approach with two FPNs inspired by the literature on optical flow estimation. 

Table-\ref{table:vton_quant} gives a quantitative comparison between the approaches discussed in this review on the VITON dataset \cite{han2018viton}. VITON was collected by crawling the web for Women, frontal view, and top clothing image pairs, and contains 14,221 pairs for training and 2,032 pairs for testing. Due to the change in metrics over the years, some of the spaces have been left empty. It can be seen that newer approaches such as SDAFN and ClothVTON+ are naturally performing better than the older approaches.

\begin{table}
\centering
\begin{tabular}{| p{2cm} | p{1cm} | p{1.5cm} | p{1.5cm} | p{3cm} | p{3cm} |} 
\hline
 \textbf{Method} & \textbf{Stages} & \textbf{3D modelling} & \textbf{Target Segmentation Map used} & \textbf{Cloth deformation method} &  \textbf{Generation}\\
 \hline
    VITON \cite{han2018viton} & 2 & \xmark & \xmark & TPS transformation with shape context matching \cite{993558} - from VITON paper & Coarse generation in stage 1 and refinement in stage 2 by $\alpha$ composition \\
    \hline
    CPVTON \cite{wang2018characteristicpreserving}, CPVTON+ \cite{Minar_CPP_2020_CVPR_Workshops} & 2 & \xmark & \xmark & Learnable TPS transformation & 2nd stage UNet provide initial prediction $I'$ along with $\alpha$ composition mask for merging $I'$ and warped clothing $C'$ \\
    \hline
    ClothFlow \cite{clothflow} & 3 & \xmark & \cmark & Two FPNs with cascaded coarse to fine dense flow estimation & 3rd stage UNet provide initial prediction $I'$ along with $\alpha$ composition mask for merging $I'$ and warped clothing $C'$ \\
    \hline 
    SDAFN \cite{sdafn} & 1 & \xmark & \xmark & Two FPNs with cascaded coarse to fine multi-dense flow estimation using DAFN & Shallow Encoder-Decoder \\
    \hline
    Cloth-VTON+ \cite{clothvton} & 4 & SMPL and SMPLifyX & \cmark & 3D cloth reconstruction and 3d cloth deformation & Parts generation UNet and Try-on fusion UNet \\
 \hline
\end{tabular}
\caption{Summary for Virtual-Try On methods}
\label{table:vton_method}
\end{table}

\begin{table}
\centering
\begin{tabular}{| p{3cm} | p{2.5cm} | p{2.5cm} | p{2.5cm} |} 
\hline
 Method & IS $\uparrow$ & SSIM $\uparrow$ & FID $\downarrow$\\
 \hline
    VITON \cite{han2018viton} 
    & 2.514 $\pm$ 0.130 & - & -\\
    \hline
    CPVTON+ \cite{Minar_CPP_2020_CVPR_Workshops} 
    & 3.1048 $\pm$ 0.106 & 0.8163 & - \\
    \hline
    ClothFlow \cite{clothflow}
    & - & 0.841 & - \\
    \hline 
    SDAFN \cite{sdafn} 
    & 2.859 & 0.888 & 10.97 \\
    \hline
    Cloth-VTON+ \cite{clothvton} 
    & - & 0.8937 & -  \\
 \hline
\end{tabular}
\caption{Quantitative results for Virtual-Try On methods}
\label{table:vton_quant}
\end{table}

\subsection{Landmark Detection}

\begin{table}[htbp]
\centering
\begin{tabular}{|p{2.5cm}|p{7cm}|p{3.5cm}|}
\hline
\textbf{Method} & \textbf{Approach \& Key Contributions} & \textbf{Limitations} \\
\hline
FashionNet \cite{fashionnet} & Simultaneously predicts clothes attributes and landmarks using a single CNN network. & Limited to a single CNN architecture and may have difficulty capturing complex spatial dependencies. \\
\hline
DFA \cite{liu2016fashion} & Three-stage CNN framework. Introduces pseudo-labels and offset predictions. & Performance heavily relies on the accuracy of the initial bounding box and may struggle with extreme pose variations. \\
\hline
DLAN \cite{yan2017unconstrained} & Estimation of landmarks without the requirement for clothing bounding boxes. Introduces joint bounding box and landmark estimation. & May face challenges in cases where clothing details are occluded or heavily distorted. \\
\hline
BCRNN \cite{attentivenet} & Predicts confidence maps for each landmark. Considers kinetic and symmetric relations between landmarks. & Performance may be impacted by complex clothing deformations and occlusions. \\
\hline
Global-Local Embedding Module \cite{lee2019globallocal} & Uses non-local operations for global feature extraction. Incorporates global features as a residual connection. & May have limitations in handling large-scale datasets due to computational complexity. \\
\hline
Spatial-Aware Non-Local Attention \cite{li2019spatialaware} & Incorporates spatial awareness using Grad-CAM in non-local attention operations. Pre-trains a ResNet-18 network on DeepFashion-C category annotations. & Spatial constraints may not generalize well to different poses, scales, and clothing styles. \\
\hline
Dual Attention \cite{9022135} & Combines spatial and channel-wise attention. Replaces the spatial attention block with a Spatial Attentive Upsampling (SAU) block. & Attention mechanisms may produce redundant attention maps and can be computationally expensive. \\
\hline
Deep Residual Spatial Attention Network \cite{9643316} & Embeds direction-aware information using Direction-Aware Spatial Attention Module. Fuses multi-level features. & Limited explanation of the direction-aware information. \\
\hline
\end{tabular}
\caption{Comparison of Fashion Landmark Detection Methods}
\label{tab:tabh}
\end{table}

Landmark detection refers to the task of identifying N fashion landmarks on a given Image I. These landmarks locate positions like  Left Collar, Right Sleeve, etc. It should be noted that the fashion/clothing landmark detection task is different and more difficult than Human Pose Estimation task. Clothes may undergo non-rigid deformations or scale variations, which is not possible for human joints because of restricted movements and positions.

Earlier works on Fashion Landmark detection worked on the assumption that clothes bounding boxes are given for images, i.e., cropped images focussing only on the cloth for which the landmark needs to be predicted is focussed. A typical dataset used was DeepFashion.  Fashion landmark detection was first introduced by \cite{fashionnet} under the same assumption of clothing boundary box. They introduced FashionNet which simultaneously predicts the clothes attributes and landmarks. It has three branches for predicting landmarks, local features, and global features.  FashionNet first predicts the Landmarks and then uses landmark predictions to pool or gate the features in the local branch to extract the local features around each predicted landmark. Then concatenate local and global features to predict clothes attributes. To be specific for landmark detection, FashionNet simply used one CNN network based on VGG-16 for predicting the landmark positions. In their next work, Liu et al \cite{liu2016fashion} proposed another framework Deep Fashion Alignment (DFA) which has three stages for Deep Convolution Neural Network. At each stage of CNN, they try to improve the prediction of landmarks. The model predicts landmark positions/offset to landmark positions and pseudo-labels at each stage. At stage-1, The CNN predicts landmarks in the space of landmark configuration. It also predicts pseudo-labels which are achieved by clustering absolute landmark positions. At stage-2, CNN takes the original image and predictions from stage-1 (landmarks and pseudo-labels) as input and predicts the offset in landmark position and pseudo-labels. At this stage, the model is learning in the space of landmark offset. The pseudo-labels here also represent the typical error patterns and their magnitude. At stage-3, the model has two branches for predicting the offset in landmarks after stage-2. To choose the branch in stage-3, the model uses error patterns predicted in stage-2. This model did not just use one CNN network to predict the landmarks and stages of the CNN model to refine the prediction in each stage. Stage-1 can be thought of as making coarse predictions, and stage-3 can be thought of as making fine predictions by predicting the offsets. The framework is similar to Residual Networks where in a way you are learning offsets to your features. 

Previous works assumed the bounding boxes on clothes which removed background clutters, and variations in human poses and scales, which are expensive to obtain and incapable in practice. Yan et al \cite{yan2017unconstrained} introduced Unconstrained Fashion Landmark Detection, where clothes bounding boxes are not provided in training and test data. They also introduced the DLAN where bounding boxes and landmarks are jointly estimated. DLAN is built upon DFA by first making it a full convolution Network and introducing two modules 1. Selective Dilated Convolutions to cope with the scale discrepancies and 2. Hierarchical Recurrent Spatial Transformer (HR-ST) to remove background clutters. Selective Dilated Convolutions are achieved by obtaining exponentially expanded receptive fields \cite{yu2016multiscale} of feature maps in layer I and applying kernel to each scale response. For HR-ST, they used the property of Spatial transformers to align feature maps for subsequent tasks. Given the feature map, the spatial transformer seeks to find the geometric transformation $\theta$ that will produce aligned feature maps. They recurrently updated geometric transformation $\theta$ in three stages to learn to remove background clutter in each stage. This geometric transformation helped them build the pseudo-boundary box they lacked in their dataset. This model was the first model to successfully handle the scale variations in the images as well as to use spatial transformers to make predictions better. 

The models that we studied until now used regression to predict landmarks however, studies in pose estimation \cite{tompson2014joint, pfister2015flowing} demonstrate this regression is highly non-linear and very difficult to learn directly, due to the fact that only one single value needs to be correctly predicted. Wang et al \cite{weng} proposed to instead of regressing landmark positions directly, we learn to predict a confidence map of positional distribution (i.e., heatmap) for each landmark, given the input image. The ground truth heatmap is obtained by adding a 2D Gaussian filter at the ground truth location of the landmark. They also introduced kinetics and symmetric relations between clothing landmarks. For example, the Left collar, Left waistline, and left Hem are connected and the Left collar is symmetrical to the right collar. It introduced Bidirectional Convolutional Recurrent Neural Network (BCRNN), which is achieved by extending classical fully connected RNNs with convolution operation for message parsing over fashion grammar. It also used Landmark-aware attention and clothing-aware attention to concentrate on functional clothing regions and to discover all the informative locations to classify diverse fashion categories and attributes respectively accurately. This paper was the first paper to use the kinetics constraints and extensively used attention modules for landmark and clothing categories. But Kinetic and symmetric constraints may not work well due to large spatial variances across poses, scales and styles of clothing items. Lee et al \cite{lee2019globallocal} addressed this issue and introduced a Global-Local Embedding Module for embedding fashion landmark information by employing a non-local block \cite{wang2018nonlocal} followed by convolution. The non-local operation is used to capture the long dependencies of two points in feature maps on each other, hence it is able to extract the global features. The features extracted from non-local are added to the original feature map as a residual connection. To capture the local features they applied two convolution operations on the output of the non-local operation. This mechanism is quite simple and works better than the constraints model. 

The previous method used a non-local block and its receptive field is the whole feature map. Because only the cloth area is the most useful area for fashion landmark detection tasks, maybe just focusing on this area might help the model to learn better representations by focusing on the target area. Keeping all these in mind Li et al \cite{li2019spatialaware} introduced Spatial-Aware Non-Local Attention. They used Gradient-weighted Class Activation Mapping (Grad-CAM) \cite{gradcam} to incorporate Spatial awareness in non-local Attention operations. They used Grad-CAM to produce spatial aware maps which will be used to attend to original feature maps in the SANL block instead of the feature map attending to itself. They pre-trained a ResNet-18 network on the DeepFashion-C category annotations and computed spatial maps for the SANL block by computing the gradients of the feature maps as weights of the feature map channels. The spatial feature basically gave the regions important for the prediction of clothes category and those are essentially the area of the clothes. 

Previous models proved that Attention models perform better than other models for landmark detection by giving more weight to the important spatial features and suppressing the less relevant positions. There have been some Graph-based models like \cite{fashion13} which gave comparable performance but SOTA models are based on attention mechanisms in the current state. The latest SOTA methods use attention in different ways. Chen et al \cite{9022135} introduced Dual attention, while previous methods only focussed on spatial attention they proposed to have channel-wise attention. They also changed the spatial attention block with Spatial Attentive Upsampling (SAU) block which integrates low-level details into final feature maps and recovers the image size and spatial details based on the non-local block and skip connections. But non-local operations may produce redundant attention maps and are computationally expensive because it generates an attention map for each element. Deep Residual Spatial Attention Network was introduced by \cite{9643316} to tackle this issue. First, they proposed Direction-Aware Spatial Attention Module to embed direction-aware information. Instead of attending the whole feature map, they attended the feature map using horizontal attention and vertical attention i.e. only in two directions. Then they proposed Spatial Attention ResBlock by integrating the DSAM with a ResBlock, i.e., input to DSAM is also added to its output after some convolution operations. They also fused multi-level features as in Feature Pyramid Network(FPN) \cite{lin2017feature}.

\subsubsection{Summary} We summarize the methods we discussed in Table-\ref{tab:tabh}. The earliest work used a single CNN to predict landmarks' locations. Then we moved to an approach that used 3-stages of CNN refining the landmark prediction at each stage. But attention-based models are now the norm. The use of attention started with the introduction of DLAN which used spatial transformation to transform the feature maps that are more aligned. The Global-Local embedding module used attention to extract global features and convolution on top of them to infuse local features. Spatial-Aware attention attended feature maps focusing on the cloth area using Grad-CAM. Dual Attention was introduced to attend channels-wise attention. Deep Residual Spatial Attention showed that attending feature maps in Horizontal and Vertical directions is enough. A few approaches took interesting approaches like BCRNN used kinetic and symmetric relations of landmarks for prediction and Dual Attention also introduced Spatial upsampling to get the same resolution as the original image.

\subsection{Attribute Recognition}
Attribute recognition in clothing refers to the task of automatically recognizing and categorizing various visual attributes or characteristics associated with clothing items. These attributes can include properties such as color, pattern, texture, style, sleeve length, neckline type, fabric type, and many others. The goal is to extract meaningful information from images of clothing items and assign labels or tags to them based on their visual features.

An early work on attribute recognition was “Describing Clothing by Semantic Attributes” by Chen et al \cite{chen2012describing}. They proposed a fully automated system that can generate a list of clothing attributes in unconstrained images. This system first performs human pose estimation to find locations of the upper torso and arms in input images, and extracts features from these regions using traditional methods like SIFT \cite{lowe2004distinctive},  maximum response filters \cite{varma2005statistical}, skin detector, etc. Each attribute has its own classifier which outputs a probability score using extracted features as input. The score reflects the confidence of attribute prediction. Finally, a Conditional Random Field (CRF) model is employed to determine stylistic relationships between attributes. 

This is one of the early methods of attribute recognition and unlike more recent methods (which will be discussed later), it does not utilize deep learning models. It uses pose estimation as a first step because estimated poses help the model to focus on parts of the input image that provide more information about clothing attributes. This is also the reason why pose estimation or landmark detection is a very common first step in models that perform attribute recognition. This step will appear again in the papers which will be discussed later. Moreover, in this method, each attribute has a dedicated SVM classifier which makes the system inefficient and not scalable for scenarios where the number of attributes is large. Instead, a multi-class classifier would better suit this classification task as it can be scaled easily with the number of attributes to be predicted. Further, by feeding the CRF model with the probability scores from the attribute classifiers, the dependencies and correlations between attributes are explored, allowing the model to retain relevant attributes and throw away incompatible ones. This makes the predictions better mimic the ground truth attributes and results in better performance than independently using the attribute classifiers.

 Owing to the success of deep learning-based methods in a variety of tasks, Liu et al \cite{fashionnet} came up with a deep learning model for attribute recognition called FashionNet which is based on Convolutional neural networks. Similar to the previous work, FashionNet first performs landmark detection on the input image to focus on informative portions of the image. Estimated landmarks are used to compute local features. These local features are fused with global image features (obtained from a CNN) to form the final feature vector which is then passed through fully connected layers to predict clothing category and attributes. 

Since deep learning models perform better than traditional approaches in a large data setup, FashionNet recognizes attributes with better accuracy than previous work. Also, unlike previous work that uses an existing pose estimation model, FashionNet has a dedicated convolutional neural network to learn to detect landmarks from data and this aids in extracting better local features.

Motivated by the success of FashionNet, Wang et al \cite{attentivenet} introduced “Attentive Fashion Grammar Network for Fashion Landmark Detection and Clothing Category Classification” (AttentiveNet). As the title suggests, they developed a fashion grammar network to improve landmark detection in FashionNet and determine clothing attributes with higher accuracy. Additionally, the model incorporates two attention mechanisms – landmark-aware attention and clothing category-driven attention – to directly improve the accuracy of attribute classification.

Landmark detection has a strong influence on attribute recognition because it tells the model which part to focus on to determine the attributes. So naturally, improvement in landmark detection leads to better attribute classification. Moreover, landmark-aware attention enables the model to interpret semantic portions of the input image. These portions provide useful information about clothing styles. But landmark-aware attention alone is not enough to perform accurate predictions. Like in FashionNet, a global feature is required to complement local information from landmark-aware attention. This is where the other attention mechanism i.e. clothing category-driven attention, comes into the picture. It propagates global information extracted from the input image and the global information combined with the local information, enhances the features to accurately predict clothing category and attributes. In contrast, FashionNet does not have any attention mechanisms and its landmark detection network is not as sophisticated as in this work. All these reasons contribute to the success of AttentiveNet.

All the models so far perform pose estimation or landmark detection explicitly, before determining attributes. But the work by Fereira et al \cite{quintino2019pose} titled “Pose Guided Attention for Multi-label Fashion Image Classification” develops a model named VSAM (visual semantic attention model) which uses pose estimation only as a guide to predict category and attributes. VSAM uses a spatial attention module (a concatenation of max and average pooling across the channel axis), to highlight informative regions in the features. These features are then regularized with ground-truth heatmaps of the relevant joints obtained by OpenPose \cite{openpose}. The regularizer uses pixel-wise L2-norm difference between the estimated and the ground-truth heatmaps which acts like another loss in addition to the cross entropy loss at the category level and the focal loss at the attribute level.

In VSAM, heatmaps from OpenPose guide the attention module to learn discriminative features from certain informative regions without abandoning other regions completely unlike previous models that perform explicit pose estimation and focus only on the regions suggested by the output of pose estimation. In a way, the regularizer in VSAM is a soft guidance from OpenPose whereas in comparison, the explicit pose estimation in previous models works like a hard supervision. Moreover, previous models are completely dependent on the output of pose estimation to determine the attributes whereas VSAM is only partially dependent on ground truth pose.

Based on the models discussed so far, the general recipe for attribute recognition consists of the following steps :-
\begin{enumerate}
    \item Pose estimation/landmark detection for soft/hard supervision of features.
    \item Extraction of features using CNN and poses estimated in the previous step.
    \item Category and attribute prediction by passing the extracted features through fully connected layers.
\end{enumerate}

There is another class of models that deviates from the general recipe stated above. Instead of performing attribute recognition as a separate task, these models focus on jointly performing both human fashion segmentation and attribute recognition. The first model in this category is FashionFormer \cite{xu2022fashionformer}, a model based on Detection Transformer (DETR) \cite{dettransformer}. FashionFormer first extracts image features for each input image using a feature extractor that contains a backbone network (Convolution Network \cite{he2016deep} or Vision Transformer \cite{liu2021swin}) with Feature Pyramid Network \cite{lin2017feature} as neck. This results in a set of multiscale features which are summed up/fused into one high-resolution feature map. The multiscale features and fused features along with special attribute queries are used to decode the output attributes of the input image (along with the segmentation masks). 

Another work in this category is a model by Tian et al \cite{tian2023detr}, titled “DETR-based Layered Clothing Segmentation and Fine-Grained Attribute Recognition”. It is similar to FashionFormer as it is also based on detection transformer and jointly models the task of segmentation and attribute recognition. The novelty of this work is a multi-layered attention module in the decoder that aggregates features from different scales and merges them together. 

By jointly modeling segmentation and attribute recognition, the attribute decoder in DETR-based models benefits from information learned by the segmentation part of the decoder model and this improves the accuracy of attribute detection. Moreover, FashionFormer and the model by Tian et al, extract multiscale features from the input image which allows the decoder to attend to information hidden in different scales. In contrast, previous models work only with a single scale, and unlike DETR-based models, they don’t have the superior learning capability of transformers. Another interesting point is that DETR-based models do not use pose estimation in their pipeline, unlike models discussed previously. And, the segmentation task in DETR-based models is quite similar to pose estimation in  previous models. This might suggest that human segmentation is better suited to facilitate attribute recognition than human pose estimation.

\subsubsection{Summary}
\begin{figure}
    \centering
    \includegraphics[height=2in]{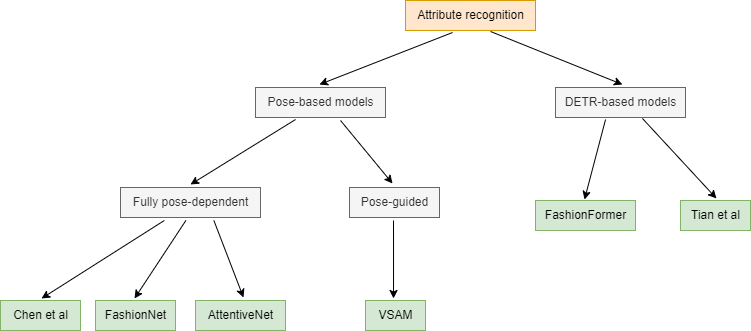}
    \caption{Categorization of models}
    \label{fig:fig1}
\end{figure}

Figure \ref{fig:fig1} shows the categorization of models for the task of attribute prediction. Models can be classified into multiple classes - fully pose-dependent, pose-guided, and DETR-based. Model by Chen et al, FashionNet, and AttentiveNet are fully pose-dependent models, VSAM is a pose-guided model, and FashionFormer and the model by Tian et al are DETR-based models.

\begin{table}[]
    \centering
    \caption{Performance Comparisons of Attribute Recognition Methods in Terms of Top-k Classification Accuracy on the DeepFashion-C dataset.}
    \resizebox{\textwidth}{!}{
    \begin{tabular}{|c|c|c|c|c|c|c|c|c|c|c|c|c|c|c|}
        \hline
        Method & \multicolumn{2}{|c|}{Category} & \multicolumn{2}{|c|}{Texture} & \multicolumn{2}{|c|}{Fabric} & \multicolumn{2}{|c|}{Shape} & \multicolumn{2}{|c|}{Part} & \multicolumn{2}{|c|}{Style} & \multicolumn{2}{|c|}{All} \\
        & top-3 & top-5 & top-3 & top-5 & top-3 & top-5 & top-3 & top-5 & top-3 & top-5 & top-3 & top-5 & top-3 & top-5\\
        \hline
        Chen et al \cite{chen2012describing} & 43.73 & 66.26 & 24.21 & 32.65 & 25.38 & 36.06 & 23.39 & 31.26 & 26.31 & 33.24 & 49.85 & 58.68 & 27.46 & 35.37\\
        \hline
        FashionNet \cite{fashionnet} & 82.58 & 90.17 & 37.46 & 49.52 & 39.30 & 49.84 & 39.47 & 48.59 & 44.13 & 54.02 & 66.43 & 73.16 & 45.52 & 54.61\\
        \hline
        AttentiveNet \cite{attentivenet} & 90.99 & 95.78 & 50.31 & 65.48 & 40.31 & 48.23 & 53.32 & 61.05 & 40.65 & 56.32 & 68.70 & 74.25 & 51.53 & 60.95\\
        \hline
        VSAM \cite{quintino2019pose} & - & - & 56.28 & 65.45 & 41.73 & 52.01 & 55.69 & 65.40 & 43.20 & 53.95 & - & - & - & -\\
        \hline
    \end{tabular}
    }
    \vspace{1em}
    \label{tab:tab1}
\end{table}

Table \ref{tab:tab1} compares the four methods analyzed above in terms of classification accuracy for different types of attributes on DeepFashion-C dataset. The huge gap between the model by Chen et al and all other models signifies the superiority of deep learning over traditional SVM classification for predicting clothing attributes. Also, AttentiveNet reports better accuracy than FashionNet supporting our insight that improvement in landmark detection leads to better attribute classification. It also confirms the role of attention in enhancing the features learned by AttentiveNet. Further, soft guidance from pose estimation in VSAM, as opposed to complete reliance on pose in other models, results in higher accuracy for some types of attributes. Overall, the low accuracy numbers for some attribute types like texture, fabric, shape, and part show that there is still a big scope for improvement in attribute prediction.

\begin{table}[]
    \centering
    \caption{Results of DETR-based models on FashionPedia dataset}
    \begin{tabular}{|c|c|c|c|}
        \hline
        \textbf{Method} & \textbf{Flops(B)} & \textbf{Params (M)} & $\mathbf{AP_{IoU + F_1}^{mask}}$\\
        \hline
        FashionFormer \cite{xu2022fashionformer} & 442.5 & 100.6 & 46.5\\
        \hline
        Tian et al \cite{tian2023detr} & 423.8 & 94.7 & 46.9\\
        \hline
    \end{tabular}    
    \label{tab:tab2}
\end{table}

Table \ref{tab:tab2} summarizes the performance of DETR-based models on the FashionPedia dataset. The last column is a joint metric to evaluate models on the joint task of segmentation and attribute prediction. It is clear from the table that there is a slight improvement in $F_1$ score as we go from FashionFormer to the model by Tian et al. But the more important observation is that the number of floating point operations and parameters in the model by Tian et al is lesser than that in FashionFormer. This is probably because Tian et al's method takes fused features and queries as inputs for the mask prediction module instead of generating mask by mask grouping and query learning like in FashionFormer. This makes Tian et al's model computationally more efficient than FashionFormer.


\section{Conclusion}\label{sec4}

This survey paper has provided valuable insights into the advancements made in Human Shape and Clothing Estimation using machine learning techniques. Specifically, we delved deeper and discussed methodologies and approaches that have demonstrated significant progress in accurately predicting human body shape, generating diverse and stylish clothing options, detecting landmarks on human bodies, and recognizing clothing attributes. The advancements discussed here have paved the way for more realistic virtual experiences, personalized fashion recommendations, and an improved understanding of clothing attributes. However, there are still challenges to overcome, such as dataset biases, scalability, and real-time performance. As the field continues to evolve, it is expected that further innovations and breakthroughs will be made, leading to even more sophisticated and accurate human shape and clothing estimation systems. These developments will undoubtedly lead to enhanced virtual experiences, personalized fashion solutions, and a deeper understanding of human shape and clothing attributes.

\printbibliography

\end{document}